\documentclass[10pt,twocolumn]{ICCAS2024}
 
\usepackage{diagbox}
\usepackage{amssymb}
\usepackage{url}
\usepackage{subcaption}

\usepackage{multirow}
\usepackage{tabularx}
\usepackage{longtable}
\usepackage{booktabs}
\usepackage{multicol}

\begin{document}

\title{Toward Integrating Semantic-aware Path Planning and Reliable Localization for UAV Operations}

\author{${}^\dagger$Thanh Nguyen Canh${}^{1,2}$, ${}^\dagger$Huy-Hoang Ngo${}^{2}$, Xiem HoangVan${}^{2}$ and ${}^{*}$Nak Young Chong${}^{1}$}

\affils{ ${}^{1}$School of Information Science, Japan Advanced Institute of Science and Technology \\
Ishikawa 923-1292, Japan (\{thanhnc, nakyoung\}@jaist.ac.jp) {\small${}^{*}$ Corresponding author}\\
${}^{2}$University of Engineering and Technology, Vietnam National University \\
Hanoi 10000, Vietnam (ngoh52180@gmail.com, xiemhoang@vnu.edu.vn) {\small${}^\dagger$ Equal contributions}\\
}

\thanks{ \noindent
  This work was supported by the Asian Office of Aerospace
Research and Development under Grant/Cooperative Agreement
Award No. FA2386-22-1-4042.
 }

\abstract{
    Localization is one of the most crucial tasks for Unmanned Aerial Vehicle systems (UAVs) directly impacting overall performance, which can be achieved with various sensors and applied to numerous tasks related to search and rescue operations, object tracking, construction, etc. However, due to the negative effects of challenging environments, UAVs may lose signals for localization. In this paper, we present an effective path-planning system leveraging semantic segmentation information to navigate around texture-less and problematic areas like lakes, oceans, and high-rise buildings using a monocular camera. We introduce a real-time semantic segmentation architecture and a novel keyframe decision pipeline to optimize image inputs based on pixel distribution, reducing processing time. A hierarchical planner based on the Dynamic Window Approach (DWA) algorithm, integrated with a cost map, is designed to facilitate efficient path planning. The system is implemented in a photo-realistic simulation environment using Unity, aligning with segmentation model parameters. Comprehensive qualitative and quantitative evaluations validate the effectiveness of our approach, showing significant improvements in the reliability and efficiency of UAV localization in challenging environments.
}


\maketitle


\section{Introduction}

In recent years, unmanned aerial vehicles (UAVs) have emerged as a significant research field and a top priority in the robotics industry, including tunnel navigation \cite{Mansouri2019}, surveillance, search operations \cite{Tomic2012}, terrain mapping \cite{MANSOURI2018}, disaster relief and accident response \cite{Jessica2018}. UAVs offer several advantages, such as simple structure and flexible flight capabilities. One of the key challenges in deploying UAVs effectively is ensuring accurate localization and navigation, particularly in complex and dynamic environments where traditional sensors like GPS and Inertial Measurement Units (IMUs) may be unreliable and unavailable. For example, IMUs often deliver suboptimal results for UAVs due to their difficulty adapting to environmental factors such as wind and air resistance. Similarly, GPS signals can be disrupted in areas with tall buildings or dense forests. The ability of UAVs to autonomously navigate and localize is therefore crucial to their operational success and safety. 

As UAV applications become more widespread, ensuring the stability of UAV self-positioning has emerged as an important concern~\cite{dai2023vision}. However, both real-time localization and mapping remain unresolved issues, because commonly used sensors can fail under adverse weather conditions~\cite{bachrach2012estimation},~\cite{costante2018exploiting}, or in the presence of mountains, tall buildings or water~\cite{bartolomei2021semantic}. In these scenarios, UAVs require reliable methods to prevent localization system failures and determine optimal flight paths to fly toward their destination. Visual Simultaneous Localization and Mapping (V-SLAM)~\cite{qian2021robust},~\cite{chen2022end}~\cite{rizk2020real} and multi-sensor fusion~\cite{canh2022multisensor} have emerged as potential solutions to address this challenge. However, these methods face significant performance issues, such as reduced robustness and accuracy, which are greatly affected by navigation environmental conditions. Typically, V-SLAM's accuracy degrades significantly in the presence of dynamic objects and areas lacking texture or with specular surfaces (\textit{e.g.} oceans and lakes). Therefore, UAV localization and navigation systems often struggle in unstructured environments due to insufficient feedback information. Consequently, integrating spatial awareness capabilities could open up the potential to improve the accuracy of UAV localization and navigation systems.

On the other hand, semantic segmentation models~\cite{zhou2022understanding}~\cite{kirillov2019panoptic}~\cite{xie2021segformer} have shown promising performance using RGB images from monocular cameras, paving the way toward integrating semantic segmentation for UAVs. Current research focuses on segmenting environmental objects, particularly utilizing 3D reconstruction to recover shapes of occluded or partially visible dynamic objects. By leveraging semantic awareness, UAVs can detect and respond to environmental factors such as terrain, moving objects, and weather conditions. This enhances self-protection and collision avoidance capabilities, while improving navigation and localization accuracy and performance.

Additionally, linking perception with SLAM ~\cite{canh2024s3m}~\cite{canh2023object} to integrate semantic information, create a semantic map, and enhance localization performance has shown the potential to address the problem of active perception. However, these methods are hindered by the high computational complexity, therefore the implementation in outdoor environments is challenging. The most similar work to ours was committed by Bartolomei \textit{et al.}~\cite{bartolomei2020perception}, proposing a method that applies semantic segmentation to enhance the localization quality of UAVs by designing a perception-aware navigation system based on the VTNet model, along with the A* Kinodynamic and B spline. In this paper, we proposed a reliable localization system for UAVs based on semantic segmentation, which integrates semantic segmentation information into a UAV path-planning framework to evaluate the quality of candidate areas for localization systems. The main contributions of this work are summarized as follows: \begin{itemize}
    \item A semantic-aware localization system for UAVs in challenging environments.
    \item A hierarchical planner that integrates the Dynamic Window Approach (DWA) algorithm with a cost map.
    \item Demonstration of the performances of our proposed system in active perception through photo-realistic simulations.
\end{itemize}

The remainder of this paper is organized as follows: Section 2 presents our proposed system based on the semantic segmentation and path planning algorithm. The experiments conducted and the analysis of the results are detailed in Section 3. Finally, Section 4 concludes the paper with discussion of future work.

\section{Methodology} \label{sec:method}

\begin{figure*}[h!ht]
\centering
\includegraphics[width=\textwidth]{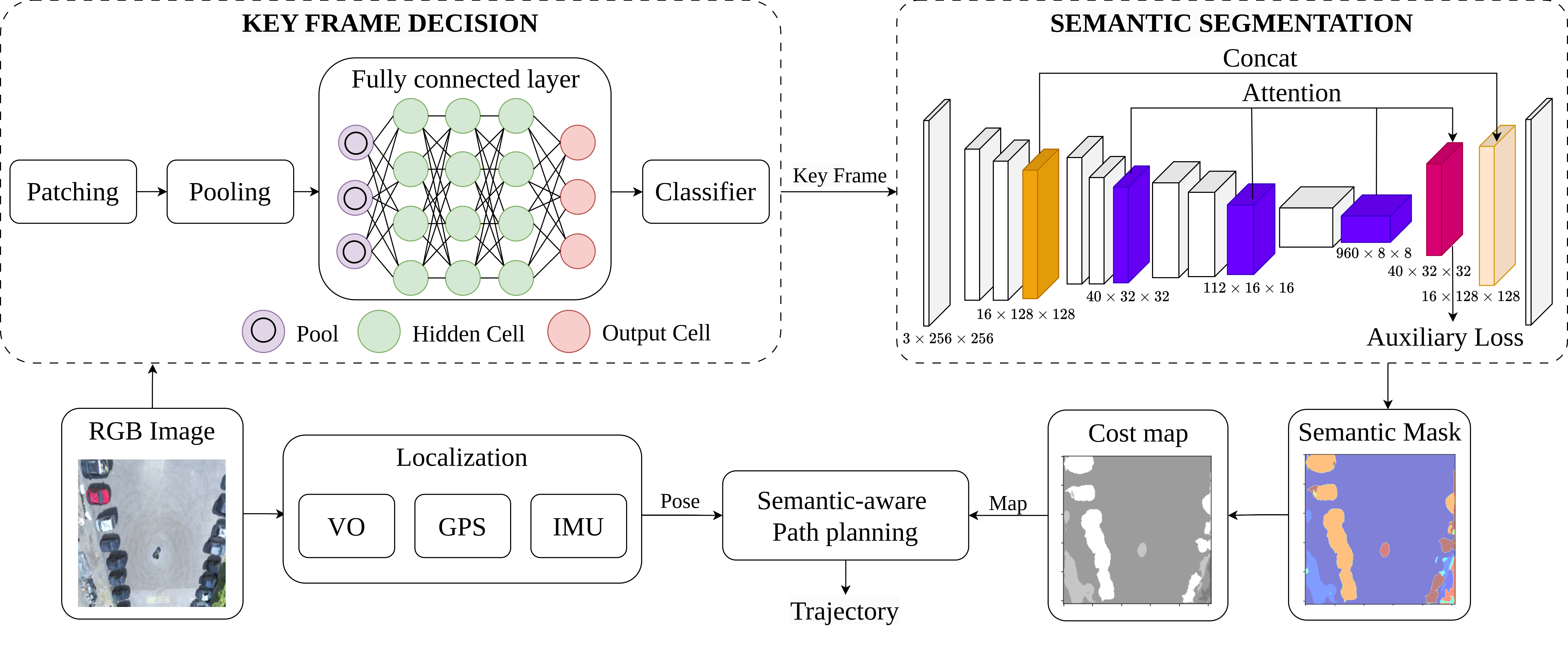}
\caption{\textbf{Overall architecture of our proposed system:} The system is composed of three main units: Key Frame Decision Module, Semantic Segmentation Module, and Integration for Semantic Information and Path Planning.}
\label{fig:overview}
\end{figure*}

Our proposed pipeline is illustrated in Fig.~\ref{fig:overview}, which takes the RGB image as input and progressively navigates to the goal based on semantic information to avoid unreliable localization areas. To achieve this, the RGB images undergo initial processing via key frame decision to determine whether semantic segmentation in this frame is necessary or not (Section~\ref{sec:keyframe} \ ~). Subsequently, we introduced an efficient semantic segmentation to extract semantic masks from individual frames and calculate the cost map (Section~\ref{sec:segmen} \ ~).  The localization module determines the UAV's pose, which can be estimated based on visual odometry, GPS, IMUs, or integration.  Finally, a hierarchical planner is performed that integrates the Dynamic Window Approach (DWA) algorithm with a cost map to provide a potential trajectory to achieve the goal and avoid unreliable localization areas (Section~\ref{sec:pathplanning} \ ~). 

\subsection{Key Frame Decision} \label{sec:keyframe}

To ensure the ability of the real-time performance of our proposed system, we first introduce a keyframe decision architecture as shown in Fig.~\ref{fig:overview} to evaluate the usefulness of RGB images. Additionally, semantic segmentation is a challenging task that may require considerable time to execute and infer results, leading to non-continuous UAV flight while the segmentation of all frames is unnecessary. Therefore, our key frame decision module determines whether the image contains sufficient or rich semantic information to be fed into the semantic segmentation module, which requires much less execution time than the semantic segmentation module.

To achieve this goal, we segment the image into distinct regions to capture comprehensive information from various areas. This segmentation is akin to the patching process utilized in contemporary vision transformer models, where the input image $\textbf{I}$ of size $H \times W$ is divided into non-overlapping smaller patches of size $P\times P$. Mathematically, this segmentation can be expressed as 
\begin{equation}
    \textbf{I} = \Big \{ \textbf{I}_{i,j}\big |i \in \big [1, \frac{H}{P} \big ], j \in \big [1, \frac{W}{P} \big ] \Big \}
\end{equation}
where $\textbf{I}_{i, j}$ denotes the patch located at the $i-th$ row and $j-th$ column of the patch grid.

Subsequently, data from each patch is globally average pooled to condense the information, with the global average pooling operation for a given patch $\textbf{I}_{i, j}$ computed as:

\begin{equation}
    GAP(\textbf{I}_{i,j}) = \frac{1}{P^2} \sum_{u = 1}^P \sum_{v=1}^{P}\textbf{I}_{i,j}(u,v)
\end{equation}
The pooled image is then flattened into a one-dimensional vector for further processing, described as $F(I) = concat \big(GAP(\textbf{I}_{1,1}, GAP(\textbf{I}_{1,2}, \cdots, GAP(\textbf{I}_{\frac{H}{P},\frac{W}{P}})\big)$. This vector is passed through a fully connected layer for evaluation, modeled by $\textbf{y} = \sigma(\textbf{W} \textbf{x} + b)$, where
$\textbf{W}$ is the weight matrix, $b$ is the bias vector, and $\sigma$ denotes the activation function. The architecture ensures meticulous assessment of pixel value distribution across image regions. By employing only a patching layer and a fully connected network, the system achieves minimal execution time, enabling real-time operation. This module is crucial for meeting the system's requirements, ensuring efficient and accurate UAV localization by leveraging the robustness of semantic segmentation and enhancing computational efficiency, making it suitable for deployment in dynamic and complex environments.

\subsection{Semantic Segmentation} \label{sec:segmen}

After selecting the keyframe, semantic segmentation plays a vital role in extracting meaningful areas of information from the surrounding environment. Prior research, including PSPNet~\cite{zhao2017pyramid} and DeepLab~\cite{chen2017deeplab} has demonstrated effective semantic segmentation using pre-trained backbone networks such as VGG16~\cite{simonyan2014very}, and ResNet~\cite{he2016deep} variants (\textit{e.g.} ResNet50, ResNet100), and MobileNet~\cite{howard2017mobilenets}. These architectures enhance model accuracy by expanding the receptive field through techniques like the Pyramid Pooling Module and Atrous Convolutions, which are employed during the encoding phase to capture broad semantic context. In the decoding phase, conditional random fields are used to smooth the output, with the aim of improving the accuracy and execution time. However, the execution time of both the DeepLab and PSPNet models remains relatively high due to their sequential nature, traversing each layer sequentially in the encoding and decoding phases. On the other hand, edge information is crucial for semantic segmentation, as it delineates boundaries between semantic regions within an image. Accurate preservation of edge information, typically found in the initial layers of backbone networks such as ResNet, InceptionNet, and VGG16, is essential for optimal model performance. Models like Unet and PsPNet prioritize edge information by using skip connections to transfer features from the encoding to the decoding module, enhancing the final segmentation accuracy. 

To address these challenges, we propose a model (Semantic Segmentation module in Fig.~\ref{fig:overview}) that supports parallel computation while leveraging edge information and maintaining rapid processing capabilities suitable for real-time systems. Our approach integrates the strengths of DeepLabv3, particularly its use of atrous convolutions, which accelerate processing and minimize the loss of critical features during pooling. Our model extracts edge information and incorporates it into the feature map at the final working layer. Additionally, features from other intermediate layers are combined with the feature map, thus requiring only two decoding layers compared to the more extensive structures in SegNet and DeepLabv3. This approach significantly improves execution time while maintaining high accuracy.

\begin{figure}
    \centering
    \includegraphics[width=0.46\textwidth]{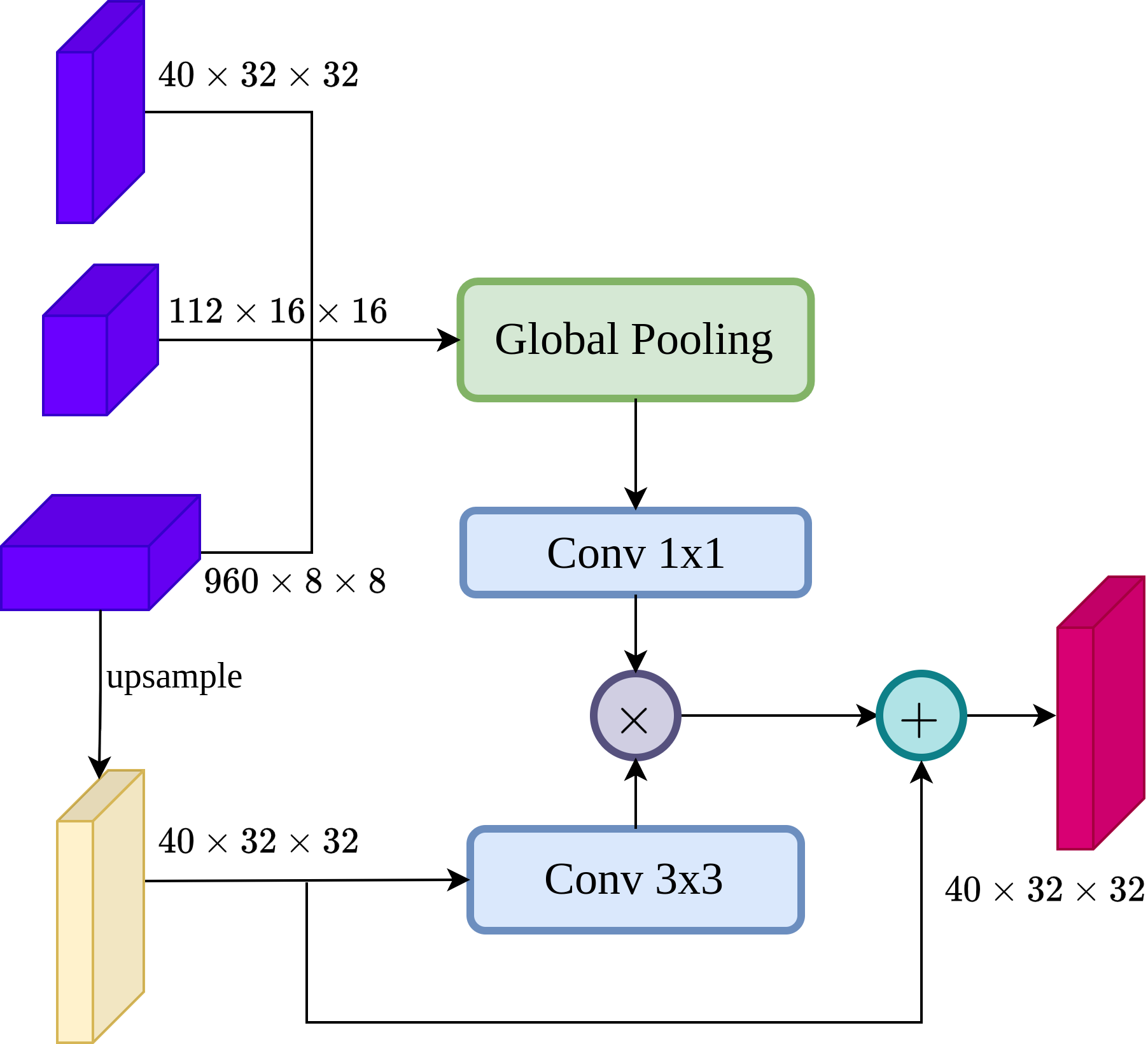}
    \caption{Attention mechanism.}
    \label{fig:attention}
\end{figure}

Moreover, the attention mechanism shown in Fig.~\ref{fig:attention} enhances the segmentation performance by focusing on the relevant parts of the images, capturing fine details and context. The attention architecture involves several steps. First, global pooling aggregates contextual information across the entire feature map:

\begin{equation}
    \textbf{g} = \frac{1}{H \times W} \sum_{i=1}^{H}\sum_{j=1}^{W} \textbf{f}_{ij}
\end{equation}
where $\textbf{g}$ is the global context vector, $H$ and $W$ are the height and width of the feature map, and $\textbf{f}_{ij}$ represents the feature vector at position $(i,j)$. 

Next, the context vector $\textbf{g}$ is passed through $1 \times 1$ convolution $\textbf{g'} = \textbf{W}_1 \textbf{g} + b_1$ with $\textbf{W}_1$ and $b_1$ is the weight matrix and bias of the $1\times 1$ convolution. The original feature map $\textbf{f}$ is processed through a  $3\times 3$ convolution to emphasize important features: $\textbf{f'} = \textbf{W}_2 * \textbf{f} + b_2$ with $\textbf{W}_2$ and $b_2$ the weight matrix and bias of the $3 \times 3$ convolution, respectively. The attention weights $\textbf{w}$ are computed as $\textbf{w} = \textbf{f'} \odot  \textbf{g'}$ with $\odot$ denoting element-wise multiplication. Finally, the refined feature map $\textbf{f}_{out}$ is obtained by adding these attention weights back to the original feature map: $\textbf{f}_{out} = \textbf{f}_{i, j} + \textbf{w}_{i, j}$.

To further boost performance, we employ an auxiliary loss technique, where the combined loss function is:

\begin{equation}
    L = \lambda \mathcal{L} + (\mathrm{1} - \lambda) \mathcal{L}_{\textit{aux}}
\end{equation}
where $\lambda$ balances the primary loss, $\mathcal{L}$ represents the discrepancy between the predicted output and the reference, and $\mathcal{L}_{\text{aux}}$ is the auxiliary loss function computed from the top layer of the decoder model.

\subsection{Semantic-aware Path Planning} \label{sec:pathplanning}

Upon receiving the semantic map, we generate a cost map based on the normalized semantic information. Since we have 23 labels, the cost map value ranges from 0 to 1 corresponding to the semantic mask value from 0 to 22. We then present a hierarchical planner that incorporates the Dynamic Window Approach (DWA)~\cite{fox1997dynamic} algorithm with this cost map. This algorithm aims to determine the optimal pair of linear and angular velocity values that describe the best achievable trajectory for the robot within the next time step. This is achieved through two main steps:
\begin{enumerate}
\item Determining feasible velocities based on constraints related to acceleration, deceleration, and obstacle avoidance.
\item Selecting the velocities that optimize an objective function.
\end{enumerate}

The objective function of the Semantic-aware DWA controller, considering the cost map, is defined as follows:

\begin{equation}
    G(v, \omega) = \alpha \cdot H(v, \omega) + \beta \cdot D(v, \omega) + \gamma \cdot Vel(v, \omega) + \epsilon \cdot C(v, \omega)
\end{equation}
where $\alpha, \beta, \gamma, \epsilon$ are positive weight coefficients, and $H(v, \omega), D(v, \omega), Vel(v, \omega)$ represent heading function, distance function and velocity function, respectively. The cost map function $C(v, \omega)$ is defined as:

\begin{equation}
    C(v, \omega) = \sum_{i=0}^k e^{-0.2t} f_k (x,y)
\end{equation}
where $t$ is the flight time and $f_k (x,y)$ is the value at the $k-th$ discrete position $(x, y)$ of the robot on the trajectory mapped onto the cost map.

Thus, the Semantic-aware DWA path planning system accounts for dynamic constraints, motion capabilities, and environmental semantic factors. The parameter $e^{-0.2t}$ indicates that initially, the UAV can choose a longer flight path to reach a better semantic area. However, as it approaches the destination, it needs to fly directly to the destination as quickly as possible, as positioning errors become less critical due to their cumulative nature.

\section{Experiments} \label{sec:results}

\begin{figure}[h!]
    \centering
    \begin{subfigure}[b]{0.24\textwidth}
    \centering
    \includegraphics[width=\textwidth]{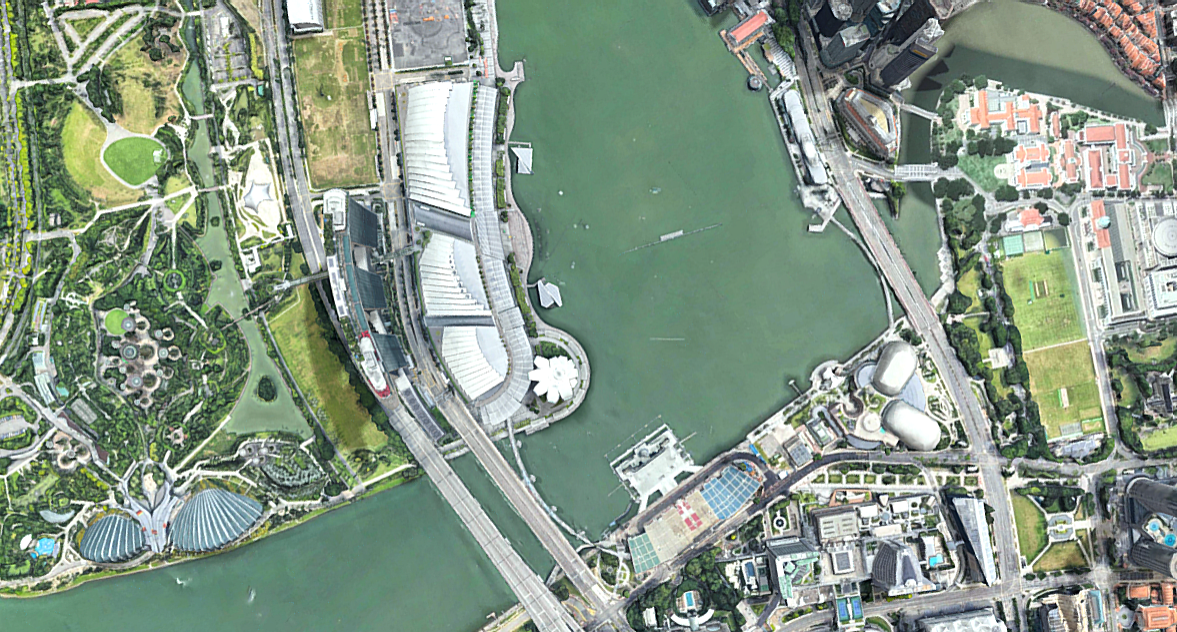}
    \caption{Singapore environment}
    \end{subfigure}
    \begin{subfigure}[b]{0.22\textwidth}
    \centering
    \includegraphics[width=\textwidth]{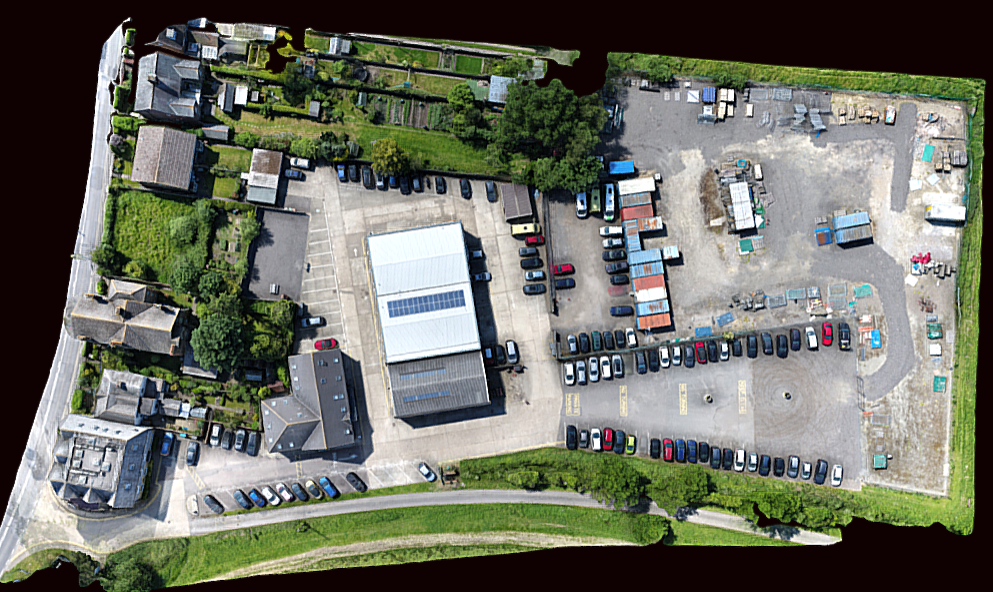}
    \caption{Baxall environment}
    \end{subfigure}
    \caption{Unity environment simulation.}
    \label{fig:env}
\end{figure}

The experimental evaluation of our proposed system was conducted in two different environments using photo-realistic Unity simulation: Baxall Village and Singapore, as depicted in Fig.~\ref{fig:env}. Creating realistic simulation environments that closely mimic real-world conditions is a crucial step in the development and testing process of our system. The Singapore Bay environment consists of four distinct areas: a grass area with low light reflection, which is a favorable area for UAV operation, and two texture-less structured areas: asphalt roads and water surfaces with strong reflections, which negatively impact UAV localization, will be considered unfavorable areas. Tall buildings in Singapore Bay disrupt the UAV's GPS signal, necessitating avoidance maneuvers. Similarly, Baxall Village features a grassy surface suitable for UAV operation, while asphalt roads, parked cars, and areas with strong reflections pose challenges for UAV flight. Finally, we utilized sockets to enable seamless communication between the Unity simulation and the semantic segmentation module.

%


\subsection{Semantic Segmentation Results}

Our proposed semantic segmentation model is trained on the UAV Image Dataset, which comprises 3,600 images captured from UAVs at altitudes ranging from low to medium (20-30m). This dataset includes 23 semantic labels that encompass crucial labels utilized in this study, such as water bodies, grasslands, cars, and high-rise buildings.

\begin{figure}[h!]
    \centering
    \begin{subfigure}[b]{0.23\textwidth}
    \centering
    \includegraphics[width=\textwidth]{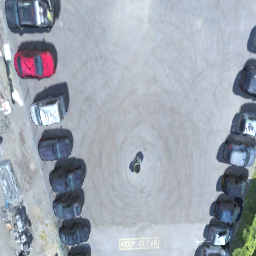}
    \caption{Label ``Yes''.}
    \label{fig:Yes_label}
    \end{subfigure}
    \begin{subfigure}[b]{0.23\textwidth}
    \centering
    \includegraphics[width=\textwidth]{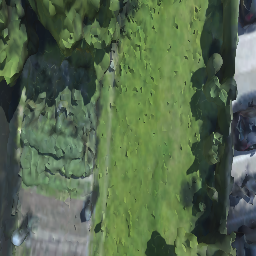}
    \caption{Label ``No''.}
    \label{fig:No_label}
    \end{subfigure}
    \caption{Example of the visual image rendered by Unity engine at training process.}
    \label{fig:pre}
\end{figure}

We evaluated the efficacy of the keyframe decision module using accuracy metrics, considering the binary classification nature of the problem. The training dataset consisted of 200 images, with 137 images labeled as ``Yes'' and 63 images labeled as ``No'', as shown in Fig.~\ref{fig:pre}. The key frame decision module achieved an accuracy of 72\% and a true positive for the ``Yes'' label reached $0.83$, demonstrating its capability to retain important cases that necessitate passing through the semantic segmentation module.

\begin{figure}[h!]
    \centering
    \begin{subfigure}[b]{0.15\textwidth}
    \centering
    \includegraphics[width=\textwidth]{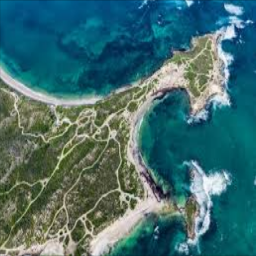}
    \label{fig:smt11}
    \end{subfigure}
    \begin{subfigure}[b]{0.15\textwidth}
    \centering
    \includegraphics[width=\textwidth]{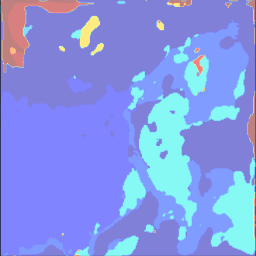}
    \label{fig:smt12}
    \end{subfigure}
    \centering
    \begin{subfigure}[b]{0.15\textwidth}
    \includegraphics[width=\textwidth]{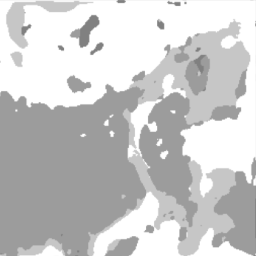}
    \label{fig:smt13}
    \end{subfigure}

    \centering
    \begin{subfigure}[b]{0.15\textwidth}
    \centering
    \includegraphics[width=\textwidth]{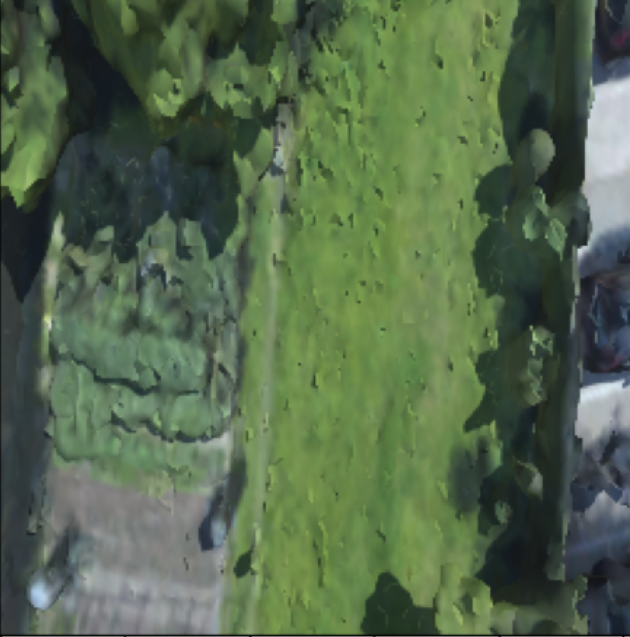}
    \caption{Input image}
    \label{fig:smt21}
    \end{subfigure}
    \begin{subfigure}[b]{0.15\textwidth}
    \centering
    \includegraphics[width=\textwidth]{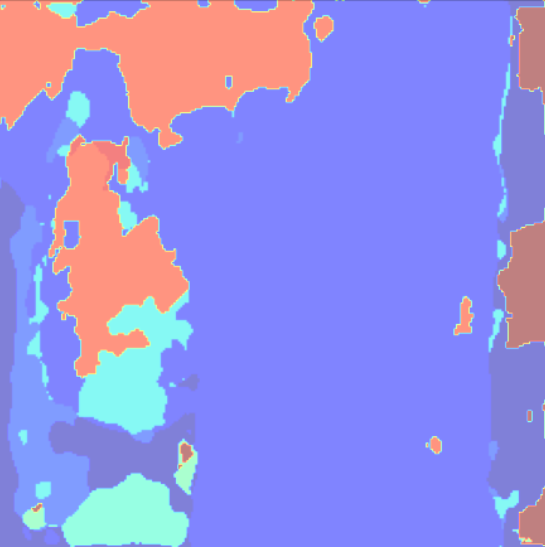}
    \caption{Semantic mask}
    \label{fig:smt22}
    \end{subfigure}
    \centering
    \begin{subfigure}[b]{0.15\textwidth}
    \includegraphics[width=\textwidth]{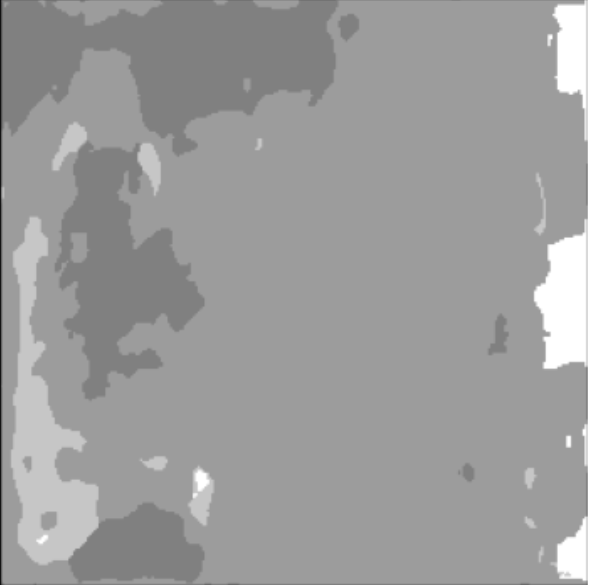}
    \caption{Cost map}
    \label{fig:smt23}
    \end{subfigure}
    \caption{Results of semantic segmentation.}
    \label{fig:sem_seg2}
\end{figure}

Table \ref{tab:1} compares our semantic segmentation model with previous models, using metrics such as mIoU (mean Intersection over Union) and average execution time (AET), averaged over 20 measurements. Our proposed model outperforms standard models like PSPNet~\cite{zhao2017pyramid}, FPN~\cite{kirillov2019panoptic}, and FAN~\cite{zhou2022understanding}. Specifically, it shows a $22.6\%$ improvement in mIoU compared to PSPNet, $21.3\%$ compared to FPN, and $15.4\%$ compared to FAN. Additionally, it enhances execution time compared to these models. While maintaining similar execution times to DeepLabv3~\cite{chen2017deeplab}, our model demonstrates an $8\%$ accuracy improvement. Compared to the Transformer model Segformer~\cite{xie2021segformer}, our model ensures a satisfactory execution time with only a slight decrease in accuracy.

\begin{table}[htbp]
\centering
\caption{Results of segmentation models.}
\begin{tabular}{p{0.15\textwidth}||p{0.12\textwidth}|p{0.12\textwidth}}
\toprule[0.5pt]
Model & mIoU & AET (ms) \\
\midrule[0.5pt]
PSPNet~\cite{zhao2017pyramid} & 0.53 & 189 \\
FPN~\cite{kirillov2019panoptic} & 0.46 & 221 \\
FAN~\cite{zhou2022understanding} & 0.50 & 101 \\
Segformer~\cite{xie2021segformer} & \textbf{0.71} & 126 \\
DeepLabv3~\cite{chen2017deeplab} & 0.59 & \textbf{91} \\
\textbf{Our method} & 0.65 & \textbf{91} \\
\bottomrule[0.5pt]
\end{tabular}
\label{tab:1}
\end{table}

\subsection{UAV Navigation Results}
\begin{figure}[htbp]
    \centering
    \begin{subfigure}[b]{0.45\textwidth}
    \centering
    \includegraphics[width=\textwidth]{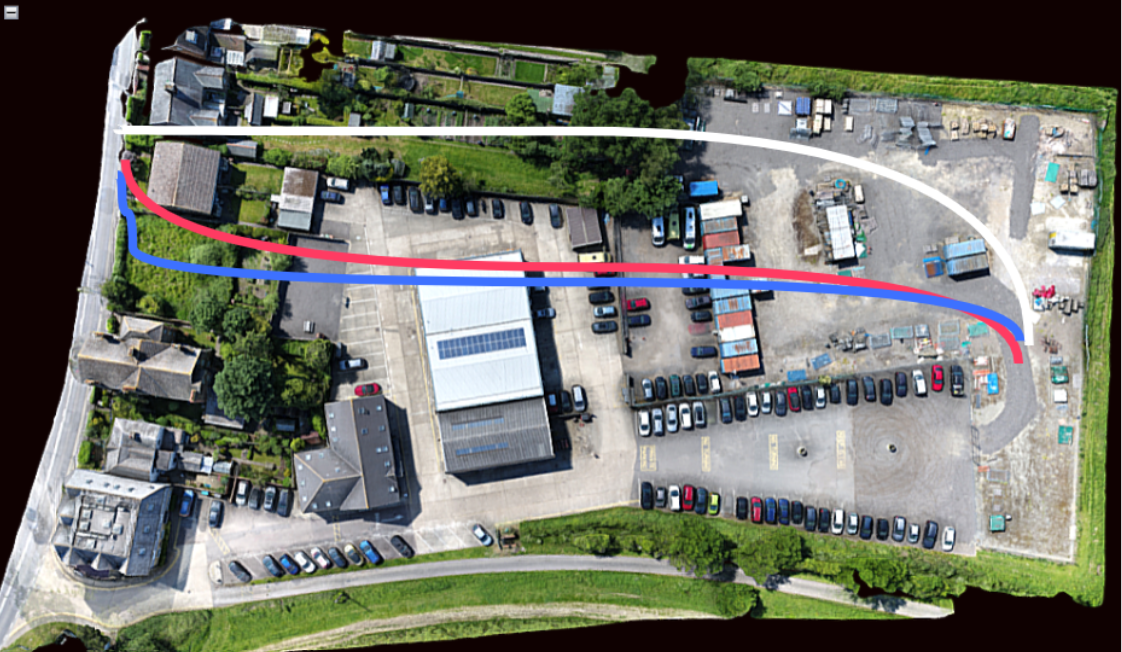}
    \caption{Baxall environment}
    \label{fig:baxall_trajectory}
    \end{subfigure}

    \begin{subfigure}[b]{0.45\textwidth}
    \centering
    \centering
    \includegraphics[width=\textwidth]{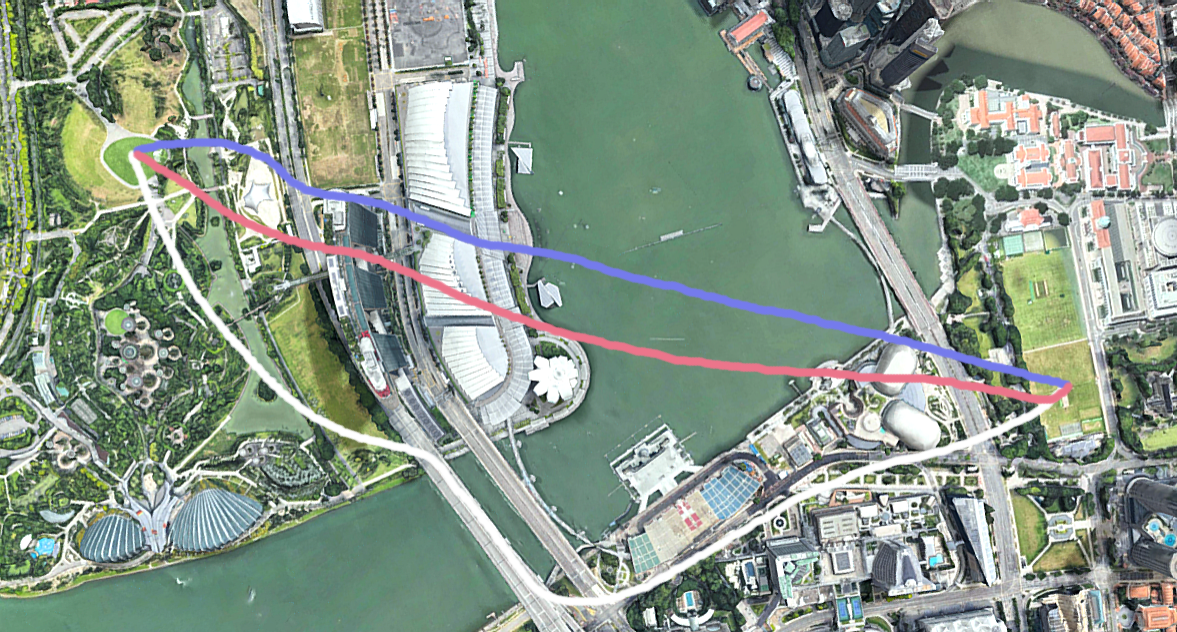}
    \caption{Singapore environment}
    \label{fig:bay_trajectory}
    \end{subfigure}
    \caption{Flight trajectories recorded: the white trajectory represents Semantic-aware DWA, the green and red trajectories represent DWA.}
    \label{fig:nav}
\end{figure}
To demonstrate the ability of navigation, we calculated the UAV's flight distance and the distance flown into areas with unreliable localization. Specifically, unreliable localized areas in the Singapore environment include roads, water bodies, and the vicinity of high-rise buildings, while in the Baxall environment, poorly localized areas are defined as roads and cars. The results are summarized in Table~\ref{tab:enhance}:

\begin{table}[ht]
  \centering
  \caption{Quantitative comparison for flight performance (Flight Distance (F-D), Unreliable Distance (U-D) meter).}
  \label{tab:enhance}
  \begin{tabular}{p{0.12\textwidth}||p{0.05\textwidth}|p{0.05\textwidth}|p{0.05\textwidth}|p{0.05\textwidth}}
     \toprule[0.5pt]
    \multirow{2}{*}{\textbf{Method}} & \multicolumn{2}{c}{Baxall} & \multicolumn{2}{c}{Singapore} \\
    & F-D & U-D & F-D & U-D \\
    \midrule[0.5pt]
    DWA~\cite{fox1997dynamic} & $17.3$ & $5.8$ & $50.6$ & $20.3$\\
    \textbf{Our method}        & $19.3$ & $3.8$ & $72.1$ & $10.2$\\
    \bottomrule[0.5pt]
  \end{tabular}
\end{table}

The flight distance of the Semantic-aware DWA system slightly climbed over the standard DWA in both test environments, increasing by $11.6\%$ in the Baxall environment and $42.5\%$ in the Singapore environment. Furthermore, the unreliable distance of the UAV's flight path shows substantial improvement, with a $52.6\%$ and $99.0\%$ increase in the Baxall and Singapore environments, respectively. This highlights the system's precision and reliability. The integration of DWA with semantic segmentation also reduces positioning errors during the initial periods, proving beneficial for UAV systems by mitigating the accumulation of errors along the flight path.

\section{Conclusion} \label{sec:conclu}

In this paper, we introduced a reliable localization system that integrates semantic segmentation with the Dynamic Window Approach (DWA) to enhance UAV navigation in texture-less and problematic areas using a monocular camera. Our proposed framework effectively addressed the challenge of navigating through areas with harder localization conditions, enabling enhanced perception of the environment for the path-planing algorithm. We demonstrated that our system can reach the assigned goal with high performance in terms of accuracy, execution time, flight distance, and unreliable distance when compared with other methodologies. Future directions could explore incorporating the uncertainty of labels in semantic segmentation and feature classifiers for UAV deployment, as well as integrating advanced machine learning techniques to further improve the UAV's perception capabilities.




\bibliographystyle{IEEEtran}
\bibliography{ref}  

\end{document}